\documentclass[letterpaper, 10 pt, conference]{ieeeconf}  % Comment this line out if you need a4paper

\IEEEoverridecommandlockouts                              % This command is only needed if 
\usepackage{graphicx} % Required for inserting images
\usepackage[dvipsnames]{xcolor}
\usepackage{framed}
\usepackage[leftmargin=4pt, vskip=8pt, font=itshape]{quoting}
\usepackage{amsfonts}
\usepackage{amsmath}
\usepackage{soul}
\usepackage[most]{tcolorbox}
\colorlet{LightLavender}{Lavender!40!}

\newcommand{\mistral}[0]{\textbf{M}}
\newcommand{\mistraltuned}[0]{\textbf{M}$_t$}
\newcommand{\phifour}[0]{\textbf{P}}
\newcommand{\phifourtuned}[0]{\textbf{P}$_t$}
\newcommand{\sonnet}[0]{\textbf{S}}

\newcommand{\vh}[0]{\textit{VirtualHome}}
\newcommand{\ttset}[0]{\textit{Task Traces}}

\newcommand{\cas}[0]{the CAS pipeline}
\newcommand{\Cas}[0]{The CAS pipeline}

\newcommand{\cmd}[2]{\texttt{\textbf{#1}(#2)}}
\newcommand{\cmdhl}[2]{\hl{\texttt{\textbf{#1}(#2)}}}

\definecolor{aliceblue}{rgb}{0.94, 0.97, 1.0}
\colorlet{shadecolor}{aliceblue}
\usepackage{lipsum}
\newenvironment{shadedquotation}
 {\begin{shaded*}\vspace{-4pt}\footnotesize
  \quoting[leftmargin=0pt, vskip=0pt, font=itshape]
 }
 {\endquoting
 \end{shaded*}
 \vspace{-4pt}
}

\tcbset{on line, 
        boxsep=4pt, left=0pt,right=0pt,top=0pt,bottom=0pt,
        colframe=white,colback=LightLavender,
        highlight math style={enhanced}
        }

\title{\LARGE \bf
Bootstrapping Human-Like Planning via LLMs
}

\author{David Porfirio$^{1}$, Vincent Hsiao$^{2}$, Morgan Fine-Morris$^{2}$, Leslie Smith$^{1}$, and Laura M. Hiatt$^{1}$% <-this % stops a space
\thanks{*This work was supported by NRL}% <-this % stops a space
\thanks{$^{1}$Navy Center for Applied Research in AI, US Naval Research Laboratory,
        Washington, DC 20375, USA
        {\tt\small \{david.j.porfirio2, leslie.n.smith20, laura.m.hiatt\}.civ@us.navy.mil}}%
\thanks{$^{2}$NRC Postdoctoral Researcher, Navy Center for Applied Research in AI, US Naval Research Laboratory,
        Washington, DC 20375, USA
        {\tt\small \{vincent.hsiao, morgan.f.fine-morris\}.ctr@us.navy.mil}}%
}

\begin{document}

\maketitle
\thispagestyle{empty}
\pagestyle{empty}

\begin{abstract}

Robot end users increasingly require accessible means of specifying tasks for robots to perform. Two common end-user programming paradigms include drag-and-drop interfaces and natural language programming. Although natural language interfaces harness an intuitive form of human communication, drag-and-drop interfaces enable users to meticulously and precisely dictate the key actions of the robot's task. In this paper, we investigate the degree to which both approaches can be combined. Specifically, we construct a large language model (LLM)-based pipeline that accepts natural language as input and produces human-like action sequences as output, specified at a level of granularity that a human would produce. We then compare these generated action sequences to another dataset of hand-specified action sequences. Although our results reveal that larger models tend to outperform smaller ones in the production of human-like action sequences, smaller models nonetheless achieve satisfactory performance.
%We also demonstrate how lightweight LLMs perform compared to larger models in several metrics.

\end{abstract}

\section{INTRODUCTION}

Robot end-user programming (EUP) tools assist novice or otherwise non-expert users in intuitively and effectively specifying tasks for robots to perform. Many such approaches design programming graphical user interfaces (such as drag-and-drop interfaces) that users can employ to correctly and fluently specify the robot's desired behavior. Such approaches give users precise control over robot behavior, but can be difficult for novices \cite{huang2017code3} or have a learning curve \cite{huang2020vipo}. Other approaches that encourage users to incompletely or abstractly specify the robot’s task, such as by specifying only a subset of task checkpoints and relying on an automated planner to come up with an actionable plan for the robot to execute, require a shift in task comprehension \cite{porfirio2024goal}. 

This problem is not unique to EUP. People in general find it difficult to generate detailed, efficient plans, even if they are meant for themselves \cite{rosenthal2020human, modi2004cmradar}. People often create physical artifacts, elaborate plans, and routines around scheduled activities to make it easier to manage the mental load of scheduling complicated goals under multiple constraints \cite{auld2011agent}.

\begin{figure}[!t]
    \centering
    \includegraphics[width=\columnwidth]{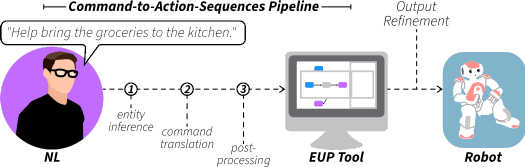}
    \caption{We envision a multi-step natural language command-to-action-sequence (CAS) pipeline as an entry point to robot end-user programming.
    }
    \label{fig:teaser}
\end{figure}

Here, we are interested in developing an approach for \textit{bootstrapping} the EUP workflow based on an initial, small set of human-provided natural language (NL) commands. 
We envision accomplishing this via a multi-step \textit{command-to-action-sequences (CAS)} pipeline that serves as the initial source of input to the EUP tool, as depicted in Figure \ref{fig:teaser}. Prior to using the tool, users provide an NL command of what they want the robot to do. Then, \cas{} uses an LLM to identify symbolic entities in the world that are relevant to the command. Next, a {\em translator Large Language Model (LLM)} translates the command and relevant entities into a symbolic representation of the key actions that the robot must perform to fulfill that command. Post-processing aligns the symbolic actions output by the LLM with the actual actions in the domain of interest. Once \cas{} has run to completion, users can then refine the action sequence via a drag-and-drop interface.

As part of this pipeline, we strongly focus on generating {\em human-like} action sequences from the NL commands. 
Mismatches in how the human \textit{expects} their NL to be interpreted by the robot and how their NL is \textit{actually} interpreted 
require explanations, which can be costly if not applied appropriately \cite{miller2017explainable}, or techniques for reconciling the human's expectation with the robot's behaviors \cite{chakraborti2019mc}. This can place a high burden on the interface and user during the refinement of the action sequence.
The production of \textit{human-like} action sequences is thus crucial to our objective, differentiating our research from prior successes in using LLMs within automated task planning \cite{song2023llm, silver2024pddl, rao2024llmplan}.

To address this, we seek to understand how well LLMs translate an NL command to a sequence of key actions that are similar to those a human would produce.
We consider several LLMs for this role, including LLMs of various sizes and LLMs that are fine-tuned on a dataset of hand-created task plans for common household tasks. 
Overall, we show that \cas{} produces human-like action sequences and how smaller LLMs perform compared to larger ones in several metrics.
Our contributions are as follows:

\begin{itemize}
    \item \textit{Technical}: a CAS pipeline that translates NL commands into \textit{human-like} action sequences.
    \item \textit{Empirical}: a characterization of how different LLMs affect the performance of the CAS pipeline.
    \item \textit{Design}: implications for integrating the CAS pipeline into future robot EUP systems.
\end{itemize}

\section{RELATED WORK}
Our research is informed by robot EUP, LLMs for task planning, and datasets of human-robot task communication. 
 
\subsection{End-User Robot Programming} 
Robot EUP has traditionally involved meticulous \textit{block-based} \cite{chung2016icustom, huang2017code3}, \textit{flow-based} \cite{glas2011IC, pot2009choregraphe}, or \textit{trigger-action} based \cite{leonardi2019tap} visual programming.  Recent research has focused on how novel interfaces can be designed to facilitate intuitive use of each of these paradigms \cite{cao2019vra, senft2021situated, ikeda2025marcer}.

Overwhelmingly, robot EUP interfaces are \textit{action-oriented} \cite{porfirio2024goal}, meaning that users express robot tasks in terms of hand-crafted \textit{task plans}, or the sequences of actions that a robot must perform in order to achieve a goal.
Hand-crafting, rather than automating, the task planning process is vital for enabling end-users to shape \textit{how} robots perform tasks. 
However, building tasks from scratch can be difficult for novice users, as evidenced by prior work \cite{porfirio2024goal, huang2017code3}.

Due to the intuitiveness of NL for human communication, \textit{natural language programming} is a promising solution for overcoming the difficulty of hand-specifying action sequences.
Historically, NL has a rich history in robot EUP, with prior work mapping NL tokens to programmatic representations \cite{ge2024cocobo, schlegel2019vajra, gorostiza2011eup, beschi2019capirci, buchina2019nli}.  
More recently, EUP systems have successfully leveraged LLMs in order to generalize beyond specific domains and alleviate the need for explicit mapping between NL and symbols \cite{karli2024alchemist, mahadevan2025imageinthat, ikeda2025marcer}.
These works excel at translating NL to individual commands, though none explore the use of LLMs for replicating hand-specified action sequences for the robot to perform. 

\subsection{LLMs for Planning}
There has been a recent surge of work that investigates how LLMs and Large Reasoning Models (LRMs) can effectively be used for planning \cite{song2023llm, rao2024llmplan, silver2024pddl}.
Despite the power of these types of models, planning remains a challenge \cite{valmeekam2024planning}, with LLM-based planners unable to meet many established planning benchmarks.
Unlike prior work that focuses on generating correct, complete, or executable plans, or plans grounded in a specific context \cite{ahn2022can}, our work has a different purpose. Informed by prior work in explainable AI that suggests humans attribute human-like psychological mechanisms to AI agents \cite{de2017people}, we instead aim to translate NL commands to action sequences in a manner similar to how a human would do it. This can then be used to pre-populate a visual EUP interface to assist the human with making further refinements to the plan. 
From an LLM planning perspective, this allows humans to correct the (messy/incorrect) plan output that LLMs produce.

\subsection{Robot Task Specification Datasets}
\label{sec:virtualhome}
Numerous datasets exist for human-robot task communication.
Of these datasets, only a few pair NL task descriptions with discretized sequences of a user's desired high-level robot actions.
As part of this work, we utilized the \textit{VirtualHome} Activity Dataset \cite{puig2018virtualhome}. This dataset contains human-generated NL descriptions of common household activities. For example, a user may describe throwing away a newspaper as: \textit{Take the newspaper on the living room table and toss it}. These descriptions are paired with human-generated ``programs'' that accomplish them, represented in a predicate-based form. There are 2821 pairs of descriptions/programs in the dataset that span numerous household tasks. 
Separate from \textit{VirtualHome}, Porfirio et al. (2023)'s \textit{Task Traces} dataset asked crowdworkers to enumerate sequences of actions that they would perform to achieve an objective and optionally pair individual actions with NL descriptions \cite{porfirio2023crowdsourcing}.
Another example is the \textit{ALFRED} dataset, which is similarly comprised of action sequences paired with NL descriptions \cite{shridhar2020alfred}. In contrast to its counterparts, however, \textit{ALFRED}'s action sequences are not created by hand. Rather, they are produced by an automated task planner.

\section{TECHNICAL APPROACH}

\begin{figure*}[!t]
    \centering
    \includegraphics[width=\textwidth]{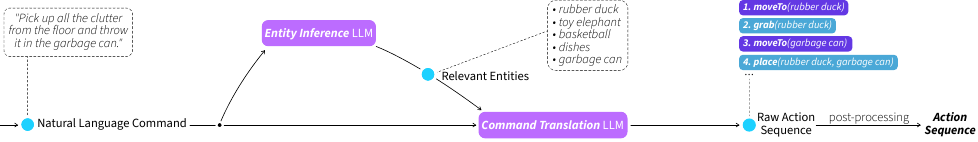}
    \caption{\Cas{} for translating NL commands into action steps.}
    \label{fig:pipeline}
\end{figure*}

The goal of this work is to use LLMs to translate between NL commands and the human-like action sequences that accomplish them. By NL command, we mean a spoken command that a person says to a robot in order to get the robot to perform desired behavior(s). It can be general (\textit{i.e.}, ``get the mail'') or specific (\textit{i.e.}, ``get the mail from the mail slot, sort it, and leave it on the kitchen counter''), depending on the user's preferences. Given this command, the robot should automatically generate a symbolic representation of how it believes it can best operationalize that behavior. We envision the user then being able to adjust the behavior in a traditional EUP interface, if necessary.

\Cas{} has three main steps to perform this translation task, each depicted in Figure \ref{fig:pipeline}. First, \textit{Entity Inference} narrows the known entities in the environment to those that are relative to the NL command being translated. Second, the \textit{Command Translation} step uses an LLM to translate the NL command and relevant entities into an action sequence. Third, a \textit{Post Processing} step maps the action sequence's commands to ones that are pertinent to the domain. We next describe each step in more detail. 

\subsection{Entity Inference}
Given the set of known entities in the robot's environment, the first step in our pipeline is to narrow this set to the entities that are pertinent to the user's command. To do this, we use off-the-shelf Mistral Codestral 22B v0.1. For example, given an NL command, ``Find your roommate and tell them they have a phone call,'' the prompt looks like: 

%\begin{shaded*}
\begin{shadedquotation}
\noindent The set of entities in the world is: master\_bedroom\_lamp, bedside\_table, desk, master\_bedroom, hallway, bathroom, car, garage, bedroom, bedroom\_lamp, refrigerator, kitchen\_cabinets, countertop, kitchen, back\_door, table, living\_room, living\_room\_lamp, entrance, coffee\_table, front\_door, living\_room\_cabinets, vacuum, clock. \\ \\
Given the command ``Find your roommate and tell them they have a phone call'', provide a short list of the entities only from this set that relate to this command.
\end{shadedquotation}
%\end{shaded*}

\sethlcolor{aliceblue}
The list of known entities can change depending on the robot's domain. Given the prompt above, the output  is \hl{\texttt{phone, roommate}}.

\subsection{Command Translation}
Next, \cas{} uses another LLM to perform command translation, where the NL command is translated into an action sequence. To perform this translation, we provide the LLM with prompts of the following form, filling in the entities and the task description as appropriate:

\begin{shadedquotation}
\noindent\textit{You are a robot. Given a task expressed in NL, you need to produce the steps necessary to achieve the goals of the task. \\ \\
The entities in the environment are as follows: phone, roommate \\ \\
Your task in NL is as follows: \\
``Find your roommate and tell them they have a phone call.''}
\end{shadedquotation}

If using a pretrained (not fine tuned) LLM for command translation, the prompt additionally requires a list of actions in predicate form that are available to the robot.
If fine-tuned on a dataset of action sequences (\textit{e.g.}, the \vh{} dataset), a list of available actions is not necessary. An example output from the above prompt is \cmdhl{Walk}{roommate}, \cmdhl{Find}{phone}, \cmdhl{Grab}{phone}, \cmdhl{TurnTo}{roommate}, \cmdhl{LookAt}{roommate}, \cmdhl{PointAt}{phone}, \cmdhl{Talk}{``I found my phone!''}, \cmdhl{PutObjBack}{phone}.

\subsection{Post-Processing}\label{sec:post}
The command translation step produces a raw action sequence (Figure \ref{fig:pipeline}, right) that requires post-processing before being given to an EUP tool or a robot.
If the command-transation LLM has been fine-tuned on a dataset like \vh{}, the actions in the raw output may differ from the actions that are recognizable by the EUP tool or the robot.
Keyword-based mapping between the dataset actions and the set of acceptable actions suffices here.
Dataset actions that fail to map to acceptable actions, refer to nonexistent entities, duplicate the previous action, or are extraneous (such as \cmd{wait}{}, which does nothing) are removed.

\section{EVALUATION}
We evaluated the effectiveness of \cas{}'s ability to produce action sequences that match human-produced action sequences.
Our evaluation considers five different LLMs for the \textit{Command Translation} step, depicted in
Table \ref{table:models}.
Our evaluation focuses on characterizing how lightweight, smaller LLMs perform this task against larger, heavyweight models, and how fine-tuned models perform against their pre-trained counterparts. 
Note that the pretrained version of Mistral 7B (\mistral{}) did not produce consistently usable output; thus, \mistral{} is excluded from further analyses.

% ID model fine-tuned parameters
\begin{table}[h!]
\small
\centering
\begin{tabular}{lcccc}
    &       & Fine- &            & Included in\\
 ID & Model & Tuned & Parameters & Further Analysis\\
 \hline

     \mistral{} & Mistral & & 7B & \\
     \mistraltuned{} & Mistral & \checkmark & 7B & \checkmark\\
     \phifour{} & Phi-4 & & 14B & \checkmark\\
     \phifourtuned & Phi-4 & \checkmark & 14B & \checkmark\\
     \sonnet & Sonnet v2 & & unk. (large) & \checkmark\\

 \hline
\end{tabular}
\caption{LLMs used in our evaluation.}
\label{table:models}
\end{table}

\subsection{Fine-Tuning \mistraltuned{} and \phifourtuned{}}

Using QLoRA \cite{dettmers2023qlora}, we fine-tuned \mistraltuned{} and \phifourtuned{} on the \vh{} dataset described in \S\ref{sec:virtualhome}, which contains 2821 hand-generated NL descriptions paired to hand-created action sequences.
\mistraltuned{} was fine-tuned with 1000 steps.
\phifourtuned{}{}, a larger model, was fine-tuned with 2000 steps.
Our training-validation split is 80\% and 20\%, respectively.

Due to being fine-tuned with the \vh{} dataset, \mistraltuned{} and \phifourtuned{} produce actions defined via \vh{} predicates.
As described in \S\ref{sec:post}, we use a keyword-based approach to map \vh{} actions to those accepted by our pipeline.

\subsection{Evaluation Dataset}\label{sec:eval_dataset}
The \vh{} dataset, while large and comprehensive for capturing how humans describe tasks using natural language, is not strongly indicative of EUP paradigms. Specifically, participants were required to specify actions sequences that included \textit{all} steps necessary for a robot to execute a task \cite{puig2018virtualhome}. In contrast, many EUP paradigms allow users to omit implied steps and let a planner fill in the small details automatically. 
To this end, we looked to a different dataset to evaluate how well this pipeline would apply to EUP tools.

\begin{figure}[!t]
    \centering
    \includegraphics[width=\columnwidth]{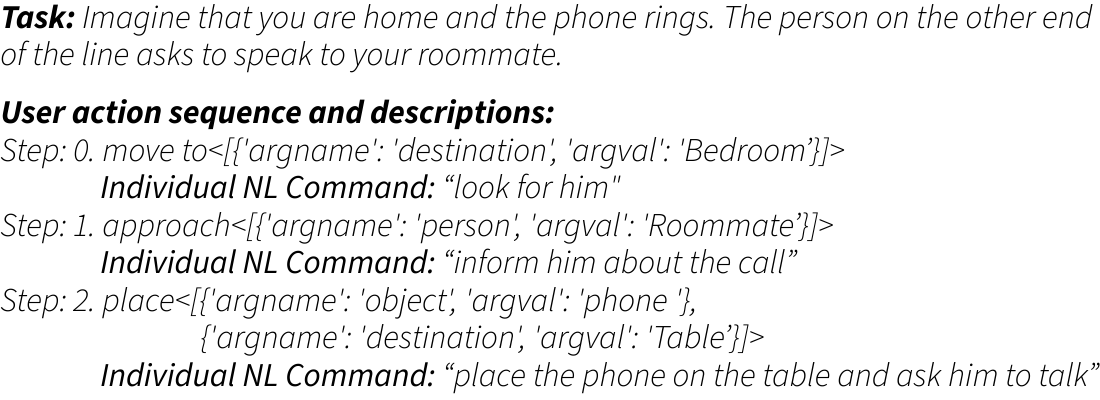}
    \caption{Simple data point from the \textit{Task Traces} dataset \cite{porfirio2023crowdsourcing}.}
    \label{fig:crowd}
\end{figure}

Evaluating \cas{} requires a dataset that pairs action sequences with a single, overarching NL description.
The dataset must also be comprehensive, that is, encompass multiple task categories.
Few datasets exist that fully satisfy these requirements, though the \ttset{} dataset \cite{porfirio2023crowdsourcing} comes close.
This dataset is comprehensive---it is comprised of 207 action sequences created by crowdworkers for 18 different task categories that someone may perform around the home, ranging from getting the mail to answering the front door.
However, the dataset does \textit{not} pair NL descriptions to \textit{whole} action sequences. 
Rather, crowdworkers optionally annotated individual actions with individual NL descriptions.
A simplified task, example action sequence, and its corresponding description sequence is shown in Figure \ref{fig:crowd}; see \cite{porfirio2023crowdsourcing} for full details of the dataset.

% eliminated 150 then another 17
Of the 207 action sequences, we first eliminated 150 that were purely social tasks or contained critical actions that lacked NL descriptions.
Determining \textit{critical} actions involved an author reviewing each action by hand to determine if its inclusion in the sequence is implied by a future action; if not, the action is critical. For example, if an action sequence is $\{\cmd{move\_to}{broom}, \cmd{grab}{broom}\}$, then \cmd{move\_to}{broom} is not critical while \cmd{grab}{broom} is.
We then eliminated another 17 sequences where the action sequence did not match the corresponding description sequence (such as if a user described putting down the groceries, but the actions specify only moving into the kitchen).
This second stage of elimination was performed by two authors coding each example separately and then meeting to resolve differences. 40 action sequences remained. 

We then augmented the \ttset{} dataset to ensure that it meets the requirement of having a single overarching NL description per whole action sequence. 
Specifically, for each action sequence, we created overarching NL summaries of the individual NL commands paired to individual actions.
Three summaries were created per action sequence.
Two of these summaries were generated by the authors of the paper, who were instructed to consider the task, the NL descriptions paired to each action in sequence, and the actions themselves.
Additionally, we used off-the-shelf Anthropic Sonnet 3.5 v2, with a temperature of 0, to generate a third overarching NL summary.
The prompt began with a description of the task similar to that in the original dataset, followed by instructions for the LLM and the NL descriptions of the action sequence: 

\begin{shadedquotation}
\noindent You are home and the phone rings. The person on the other end of the line asks to speak to your roommate. Please summarize the following steps as a short sentence. Skip or combine unimportant steps. Phrase it as a command to yourself in the second person.  \\ \\
1. look for him  \\
2. inform him about the call \\
3. place the phone on the table and ask him to talk 
\end{shadedquotation}

\noindent An example human-generated summary of these steps is \hl{\textit{Inform my roommate of the call and place the phone on the table for him.}} Example LLM-generated output is \hl{\textit{Find your roommate and tell them they have a phone call.}}

\subsection{Evaluation Measures}\label{sec:measures}

\begin{figure*}[!t]
    \centering
    \includegraphics[width=\textwidth]{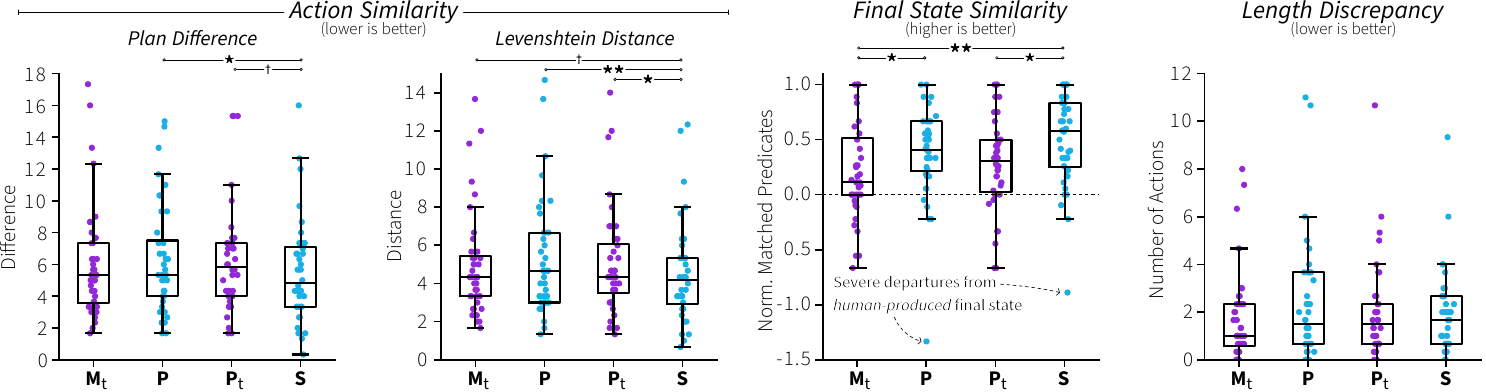}
    \caption{Results for \textit{Action Similarity} (left), \textit{Final State Similarity} (center), and \textit{Length Discrepancy} (right). \dag, *, and ** denote $\mathrm{p}<0.1$, $\mathrm{p}<0.05$, and $\mathrm{p}<0.01$, respectively.}
    \label{fig:results}
\end{figure*}

Our evaluation includes several measures intended to compare LLM-generated action sequences to their human-produced counterparts.
We consider action sequences to be human-like if they (1) contain the same actions as their human-created counterparts (\textit{action similarity}); (2) are similar to their human-created counterparts in the final state of the world resulting from executing the sequences (\textit{final state similarity}); and (3) are substantially different in length to their human-created counterparts (\textit{length discrepancy}).

First, we consider \textit{action similarity} via two separate measures, the first being \textit{plan difference}, which is a measure of distance between two action sequences that focuses on actions that are different between the two sequences, while de-emphasizing the order in which those actions appear \cite{fox2006plan}. The distance is calculated as $|A-B| + |B-A|$ for two action sequences A and B, or the number of actions that appear in one action sequence but not the other (and vice versa) without regard for action order. Lower \textit{plan difference} indicates higher \textit{action similarity}.

Our second \textit{action similarity} measure considers the order in which actions appear, and is the \textit{Levenshtein distance} between the two action sequences. This distance calculates the number of edits that must be made to an action sequence $A$ in order to turn it into action sequence $B$, including insertion, deletion and replacement edits. Lower \textit{Levenshtein distance} indicates higher \textit{action similarity}. 

For our next measure, \textit{final state similarity}, we calculate the similarity between the final state of the world produced by LLM-generated action sequences and their human-created counterparts.
We first translate each action sequence to the \textit{Interaction Specification Language (ISL)} \cite{porfirio2025ISL}, which requires a planning problem definition. We based our problem definition on the \ttset{} dataset.
We then calculate an actionable plan for each sequence using Porfirio et al. (2024) \cite{porfirio2024goal}.
Actions deemed impossible for the robot to perform, such as \cmd{grab}{car}, are ignored.

Then, to perform the similarity calculation, let $I$ be the initial state of the world, $F_h$ be the human-produced final state, and $F_l$ be the LLM-produced final state.
The similarity calculation is the size of the difference between the LLM-produced and human-produced states normalized by the size of the difference between the human-produced state and the initial state. We then subtract this value by 1 to create an inverse relationship, where higher similarity values are better: $1 - |F_h - F_l|\div|F_h - I|$.

For our final measure, \textit{length discrepancy}, we consider the difference in length between LLM-produced action sequences and their human-created counterparts.
To calculate this measure, we count the actions in each action sequence and take the absolute value of the difference between its length and the length of its human-created counterpart.
Note that our measures for \textit{action similarity} implicitly account for length differences.
However, explicitly calculating the difference in lengths helps determine how much \textit{action similarity} can be accounted for by varying plan lengths.

\subsection{Evaluation Procedure}

First, we fed the NL summaries created in \S\ref{sec:eval_dataset} to \cas{}.
Next, we compared each output from \cas{} to its corresponding human-generated action sequence using the measures discussed in \S\ref{sec:measures}. 
Recall that for each action sequence, there are three NL summaries---two originating from the authors and one originating from an LLM.
For each of our measures, we averaged the output from each of the three summaries to produce a single value.

\section{RESULTS}

\paragraph{Action Similarity}
A Friedman test detected a marginal effect of LLM model on \textit{plan difference},  (Figure \ref{fig:results}, left), $\chi^2(3)=6.75$, p $= 0.080$. Post-hoc Wilcoxon tests with the Bonferroni correction\footnote{We applied the Bonferroni correction by multiplying the p-value of each pairwise comparison by the total number of comparisons, which is six.} indicate a significant difference between \phifour{} and \sonnet{} (p $=0.024$) and a marginal difference between \phifourtuned{} and \sonnet{} (p$=0.083$).
Another Friedman test detected a significant effect of LLM model on \textit{Levenshtein distance} (Figure \ref{fig:results}, center left), $\chi^2(3)=7.93$, p $= 0.048$, with pairwise differences between \phifour{} and \sonnet{} (p $<0.01$) and \phifourtuned{} and \sonnet{} (p $=0.036$), and a marginal difference between \mistraltuned{} and \sonnet{} (p $=0.071$). \textbf{\textit{Takeaway:} The largest pretrained model (\sonnet{}) tends to produce better-matching action sequences than the smaller models.}

\paragraph{Final State Similarity}
A Friedman test detected a significant effect of LLM model on \textit{Final State Similarity} (Figure \ref{fig:results}, center right), $\chi^2(3)=21.8$, p $< 0.001$.
Post hoc comparisons with the Bonferroni correction indicate significance between \mistraltuned{} and \phifour{} (p $=0.011$), \mistraltuned{} and \sonnet{} (p $< 0.01$), and \phifourtuned{} and \sonnet{} (p $= 0.020$).
One-sample Wilcoxon tests indicate that the action sequences resulting from all four models produce final states that differ significantly from $0$, indicating some progression from the initial state of the world (p $< 0.01$ for \mistraltuned{} and p $< 0.001$ for \phifour{}, \phifourtuned{}, and \sonnet{}).
\textbf{\textit{Takeaway:} All models produce action sequences that are useful, advancing the robot towards ground-truth final state. Although fine-tuning helps Mistral 7B (\mistraltuned{}) produce useful action sequences compared to the incoherent output of its pretrained baseline (\mistral{}), fine-tuning does not have a significant effect on Phi-4 (\phifour{} vs. \phifourtuned{}). Larger models tend to outperform smaller models.}

\paragraph{Length Discrepancy}
We did not detect any significant effect of LLM model on \textit{length discrepancy} (Figure \ref{fig:results}, right).
Post-hoc tests revealed no significance in any pairwise comparison.
\textbf{\textit{Takeaway:} Our models produce action sequences that consistently differ in length.}

\section{CONCLUSION}

In this paper, we introduce a novel pipeline for translating natural language commands to human-like action sequences.
Our aim is to produce human-like action sequences given only natural language as input, in which \textit{human-likeness} of an LLM-produced action sequence is defined in terms of (1) similarity to its hand-crafted counterpart; (2) the degree to which the LLM-produced action sequence achieves the final state of the world achieved by its hand-crafted counterpart; and (3) the discrepancy in action sequence length to its hand-crafted counterpart.
In what follows, we summarize our findings and produce concrete implications for applying \cas{} in robot EUP. 

\subsection{Implications for Human-Like Action Sequences}
First, as evidenced by our action-sequence similarity and final state similarity results, larger LLMs appear better than smaller ones at producing human-like action sequences.
That being said, smaller models do remarkably well.
\mistraltuned{} and \phifourtuned{}, both fine-tuned small models, show no significant difference in length to their largest counterpart, \sonnet{}, and \mistraltuned{} does not show a significant difference (p $<0.05$) to \sonnet{} in action similarity.
Furthermore, all models, both small and large, produce action sequences that significantly advance the robot towards its intended final state.
\textit{\textbf{Implication:} \Cas{} should use larger models when available and practical. If using larger models is infeasible, such as when tasking robots in the field, smaller models may produce similar performance to the large ones across some metrics.}

Second, in all of our measures, fine-tuned models exhibit no significant difference to their pretrained counterparts.
However, the fine-tuned version of Mistral 7B, \mistraltuned{}, was able to produce coherent, useful output, even when its pretrained counterpart (\mistral{}) could not.
\textit{\textbf{Implication:} Fine-tuning may or may not be necessary, depending on the model being used.}

Lastly, we note that the discrepancy of lengths between human-produced and LLM-produced action sequences is largely consistent across all four models, with no significant difference between models.
\textit{\textbf{Implication:} Length discrepancy is a weak differentiator of human-likeness for the models included in our study, though future work is needed to see if it is a better differentiator for other models.}

\subsection{Limitations and Future Work}
Our future work is informed by the current limitations of \cas{}.
First, although we have shown how different LLMs cause varying performance in \cas{}, we have not yet determined what level of human-likeness is sufficient for actual human users.
Future work must therefore evaluate \cas{} with real human users in order to determine acceptable levels of action similarity, final state similarity, and length discrepancy.

Second, we have not yet embedded \cas{} into an actual EUP pipeline.
Future work must therefore connect \cas{} with an actual EUP tool, such as \textit{Polaris} \cite{porfirio2024goal}, in order to fully realize the vision of our system.
We must also test the effectiveness of \cas{} within a full-fledged EUP workflow via additional user studies.

Third, \cas{} itself still has room for improvement.
Although we tested our pipeline with LLMs that have had a subset of their weights updated via QLoRA, we have yet to test it with LLMs that have had all of their weights updated via full fine tuning.
Additionally, we have yet to determine whether fine-tuning on different datasets (\textit{e.g.,} ALFRED \cite{shridhar2020alfred}) affects model performance.
Future work must thereby investigate how different datasets and fine-tuning paradigms affect the performance of \cas{}. 

Lastly, while investigated in the context of end-user programming, our pipeline may be useful to a variety of other applications, including scheduling assistance.
Future work must investigate \cas{} in these other contexts.

\bibliographystyle{IEEEtran}
\bibliography{main.bib}

\end{document}